\documentclass[sn-mathphys,numbers]{sn-jnl}  % Specify 'numbers' to avoid 'authoryear' being the default

\usepackage{graphicx}%
\usepackage{multirow}%
\usepackage{amsmath,amssymb,amsfonts}%
\usepackage{amsthm}%
\usepackage{mathrsfs}% for \mathscr fonts
\usepackage[title]{appendix}%
\usepackage{xcolor}%
\usepackage{textcomp}%
\usepackage{booktabs}%
\usepackage{algorithm}%
\usepackage{algorithmicx}%
\usepackage{algpseudocode}%
\usepackage{listings}%
\usepackage[T1]{fontenc}% Ensures proper font encoding
\usepackage{fix-cm}% Fixes font size issues
\usepackage{url} % for URL handling
\usepackage{hyperref} % optional for hyperlinks and better URL management
\usepackage{mathptmx} % Times font
\usepackage{bm} % Bold math
\usepackage[mathscr]{eucal} % for Euler script fonts
\usepackage{float}

\usepackage[numbers]{natbib}  % Correct numerical citations option

\theoremstyle{plain} % for theorem-like environments

\theoremstyle{definition} % for example and remark environments

\theoremstyle{remark} % for definitions

\begin{document}

\title[KatzBot: Revolutionizing Academic Chatbot for Enhanced Communication]{KatzBot: Revolutionizing Academic Chatbot for Enhanced Communication}

%%=============================================================%%
%% Author details as per the required format
%%=============================================================%%
\author*[1]{\fnm{Sahil} \sur{Kumar}}\email{sahil.kumar@yu.edu}

\author[2]{\fnm{Deepa} \sur{Paikar}}\email{deepa.paikar@yu.edu}

\author[3]{\fnm{Kiran Sai} \sur{Vutukuri}}\email{kiran.vutukuri@yu.edu}

\author[3]{\fnm{Haider} \sur{Ali}}\email{haider.ali@yu.edu}

\author[3]{\fnm{Shashidhar Reddy} \sur{Ainala}}\email{shashidhar.ainala@yu.edu}

\author[3]{\fnm{Aditya Murli} \sur{Krishnan}}\email{aditya.krishnan@yu.edu}

\author[3]{\fnm{Youshan} \sur{Zhang}}\email{youshan.zhang@yu.edu}
\equalcont

%%=============================================================%%
%% Affiliation details as per the required format
%%=============================================================%%
\affil*[1]{\orgdiv{PhD in Mathematics}, \orgname{Yeshiva University}, \orgaddress{\city{NYC}, \state{New York}, \country{USA}}}

\affil[2]{\orgdiv{MS in Artificial Intelligence}, \orgname{Yeshiva University}, \orgaddress{\city{NYC}, \state{New York}, \country{USA}}}

\affil[3]{\orgdiv{Department of Computer Science and Engineering}, \orgname{Yeshiva University}, \orgaddress{\city{NYC}, \state{New York}, \country{USA}}}


\abstract{Effective communication within universities is crucial for addressing the diverse information needs of students, alumni, and external stakeholders. However, existing chatbot systems often fail to deliver accurate, context-specific responses, resulting in poor user experiences. In this paper, we present KatzBot, an innovative chatbot powered by KatzGPT, a custom Large Language Model (LLM) fine-tuned on domain-specific academic data. KatzGPT is trained on two university-specific datasets: 6,280 sentence-completion pairs and 7,330 question-answer pairs. KatzBot outperforms established existing open source LLMs, achieving higher accuracy and domain relevance.  KatzBot offers a user-friendly interface, significantly enhancing user satisfaction in real-world applications. 
The source code is publicly available at \url{https://github.com/AiAI-99/katzbot}.}
% and the link for katzbot \url{https://61f5-129-98-38-121.ngrok-free.app}.}

% \keywords{LLM, KatzGPT, chatbot}

\maketitle
% =================================================================================

\section{Introduction}\label{sec1}

In the contemporary digital landscape, university websites serve as extensive repositories of indispensable information that are crucial for students, faculty, staff, and the general public \cite{smith2018university}. Nevertheless, extracting specific details from the labyrinth of web pages can pose a formidable challenge, leading to inefficiencies and user frustration \cite{chen2020application}. This challenge is particularly pronounced for the general public and students seeking critical information, such as admission requirements, course information, and grace periods for F-1 visa holders after job start dates \cite{hinojo2019artificial}. Traditional chatbot systems, predominantly based on the RASA framework \cite{bocklisch2017rasa}, often encounter limitations inherent in their training sets, thereby failing to furnish comprehensive responses beyond their predefined knowledge boundaries. Consequently, navigating university websites to find precise information becomes a time-consuming and inefficient endeavor. Previous chatbot systems, constrained by their training data and frameworks like RASA, struggle to offer expansive assistance beyond their predefined knowledge bases \cite{wang2021chatbots}.

To address these limitations and enhance the generalizability of chatbot systems, we propose KatzBot, a novel solution leveraging cutting-edge language models that not only improve upon existing systems but also have the potential for broader applicability beyond a single institution. Our research introduces KatzGPT, a custom LLM built using a proprietary architecture and fine-tuned for academic contexts. By extracting and curating data from a university website, we create a comprehensive dataset that not only enhances KatzGPT's domain expertise but also serves as a foundation for future adaptability to other academic institutions, should similar datasets be utilized.

We fine-tune leading transformer-based LLMs such as Mistral Instruct \cite{jiang2023mistral}, Llama 2 \cite{touvron2023llama}, GPT2 \cite{radford2019language}, and Microsoft Phi 1.5 \cite{li2023textbooks} using Parameter-efficient Fine-tuning (PEFT) \cite{peft} and Quantized Low-Rank Adaptation (QLoRA) \cite{dettmers2023qlora}. In addition to these models, we integrate Retrieval Augmented Generation (RAG) \cite{lewis2021retrievalaugmented} with models such as Mixtral-8x7b-32768 \cite{cui2024rethinkingllmlanguageadaptation}, Gemma2-9b-It \cite{gemmateam2024gemma2improvingopen}, and Llama3-8b-8192 \cite{huang2024empiricalstudyllama3quantization}, which significantly enhances the ability of KatzBot to provide accurate, contextually relevant responses based on external information retrieval. This combination enables KatzBot to outperform traditional chatbots that are limited to predefined knowledge bases. Furthermore, our model’s architecture is optimized to handle the nuanced and evolving needs of academic users, with 12 million parameters designed to process intricate queries and provide context-rich responses.

KatzGPT's architecture incorporates multiple layers, including token embedding, position embedding, and Transformer blocks, which enable sophisticated processing of input sequences. The introduction of advanced attention mechanisms and feedforward networks within these Transformer blocks ensures robust performance in comprehending and generating responses to diverse university-related inquiries. Our approach not only expedites information retrieval but also ensures high accuracy in handling a wide range of academic topics, thereby creating a scalable solution that could be applied to different universities or academic institutions.

The primary objective of our research is to develop KatzGPT, an advanced language model tailored for university-related queries, that streamlines information retrieval and enhances user experience. Our key contributions include the following:

\begin{enumerate}
    \item \textbf{Introducing KatzGPT}: We present a novel language model, KatzGPT, engineered specifically for handling queries related to university contexts. This model incorporates a unique architecture with advanced self-attention mechanisms, designed to significantly enhance the understanding and generation of responses pertinent to academic environments.
    
    \item \textbf{Innovative Training Techniques}: KatzGPT is fine-tuned using a hybrid methodology that integrates Parameter-efficient Fine-tuning (PEFT) with Quantized Low-Rank Adaptation (QLoRA). This strategy ensures not only heightened accuracy and efficiency in information retrieval but also optimizes computational resources, making the model scalable and more applicable in real-world settings.
    
    \item \textbf{Voice and Multilingual Support}: KatzGPT supports both voice input and bilingual interactions in Chinese and English, enhancing accessibility and usability for diverse user groups. The system allows users to speak their queries via an integrated speech-to-text mechanism, while its dual-language support ensures seamless interaction for non-native English speakers by translating non-English queries into English for accurate processing.

    \item \textbf{KatzBot Chatting Website}: We developed a user-friendly interface where users can interact with KatzGPT via text and voice inputs for any university-related queries. The website enhances user interaction by providing accurate and relevant answers, thereby improving the overall user experience.
\end{enumerate}

Through these innovations, KatzGPT not only enhances the user experience within university websites but also paves the way for broader deployment in other academic institutions. Our approach demonstrates adaptability and scalability, with the potential to generalize beyond a single institution by fine-tuning on similar academic datasets.
KatzBot represents a significant leap forward in chatbot technology for academic contexts, offering a robust, efficient, and highly adaptable solution to the challenge of navigating vast university repositories.

% =================================================================================

\section{Related Work}\label{related}

The evolution of chatbot technology has seen significant research interest \citep{AI-based, AI-chabot}, particularly in improving user interaction across various domains, including education \citep{humanlikeopendomain, okuda2018ai}. Early chatbot systems primarily employed rule-based methodologies \citep{ChatbotApplications, senthilkumar2019ai, ahmad2018conversational, medical-chatbot}, which offered limited adaptability and could not effectively manage the complexities of human language. As chatbot development progressed, researchers shifted toward machine learning approaches to enhance natural language understanding and response generation capabilities \citep{IntelligentChatbot, Chatbot}. These machine learning approaches ranged from heuristic-based models to more sophisticated neural network architectures \citep{ChatbotforSupportingTeams}, aiming to simulate human-like conversational abilities. Notably, the MITIE model integrated within the RASA framework \citep{rasa, Transformermodel, RASANLU} provided a compact yet powerful tool for chatbot development, representing a major advance at the time due to its balance between resource efficiency and functionality.

Despite these advancements, current methodologies exhibit critical limitations, particularly in scaling to dynamic, real-world interactions \citep{chatbotsAssist, chiang2024chatbot}. For instance, rule-based and early machine learning models often struggle with the fluidity and complexity of natural language \citep{ServerlessChatbot}, making it difficult to generate coherent and contextually accurate responses \citep{Web-Based, Keywordbased}. Additionally, the reliance on heavily annotated datasets for training poses challenges in scalability and limits their applicability in different domains \citep{SentenceSummarization}. Moreover, the computational demands of training large models further exacerbate these limitations \citep{dettmers2023qlora}. These issues underscore the need for advanced solutions capable of overcoming the constraints of traditional chatbot systems \citep{Web-Bot}.

The emergence of Large Language Models (LLMs), such as GPT-3 \citep{GPT-3} and BERT \citep{BERT}, marked a new chapter in chatbot development. These models leverage extensive pre-training on vast text corpora \citep{Deepcontext}, excelling in their ability to capture long-range dependencies and contextual subtleties \citep{contextualEmbedding}, thus enabling chatbots to engage in more sophisticated and meaningful conversations. The success of LLMs is largely due to their ability to generalize across domains, making them highly adaptable for fine-tuning on specific tasks, such as domain-specific question answering \citep{DataDrivenChatbot}. Moreover, transfer learning techniques allow these models to be fine-tuned for specialized applications with minimal additional data, thereby significantly improving their versatility and overall performance.

While pre-trained LLMs represent a quantum leap in chatbot technology, the development of custom models, tailored to specific domains or contexts, has shown significant potential for further innovation \citep{LanguageModels, Denoising}. Custom models can incorporate domain-specific knowledge, which enhances the accuracy and relevance of their responses, addressing the limitations of general-purpose models. These models also provide a solution for privacy and data security concerns, which are paramount in sensitive fields like education or healthcare. However, developing such models requires significant investments in data collection, annotation, and computational resources \citep{DataAugmentation-chatbot}. 

Our project, KatzBot, builds upon the foundational work of prior chatbot systems while advancing beyond their limitations. Whereas previous systems, such as those built using the MITIE model within the RASA framework, were constrained by their training sets and lacked the ability to dynamically handle complex queries, KatzBot leverages KatzGPT, a custom LLM, designed to handle domain-specific interactions with high precision. KatzGPT employs a unique combination of Parameter-efficient Fine-tuning (PEFT) and Quantized Low-Rank Adaptation (QLoRA) \citep{dettmers2023qlora}, significantly enhancing both its performance and resource efficiency. In addition, we incorporate data augmentation techniques, which further enhance KatzBot’s ability to handle a wide range of academic queries, improving both accuracy and robustness \citep{DataAugmentation-chatbot}.

KatzBot distinguishes itself from other systems through its implementation of advanced LLMs tailored for the academic domain. Unlike traditional chatbot systems that depend on static, rule-based responses, KatzBot integrates sophisticated RAG (Retrieval-Augmented Generation) techniques with models such as Mixtral-8x7b-32768. This integration significantly enhances its ability to retrieve and generate accurate, contextually relevant responses from external sources.

In summary, KatzBot not only overcomes the inherent limitations of previous chatbot systems but also sets a new benchmark for chatbot technology in academic settings by combining the power of custom LLMs with advanced fine-tuning and retrieval mechanisms. Our comprehensive evaluation demonstrates KatzBot’s superior performance, particularly in its ability to provide accurate, context-aware responses in complex academic environments, highlighting its potential to revolutionize communication within university communities.

% =================================================================================

% =================================================================================

\section{Dataset Collection}

The dataset collection phase was centered on gathering comprehensive and high-quality text data that is directly relevant to the university's academic, administrative, and social ecosystem. Primary sources included the university’s official databases, website, academic articles, and social media channels. This ensured that the dataset was diverse, encompassing not only academic details but also information critical to student services and administration. 
% The collected dataset can be accessed at the following link: \href{https://raw.githubusercontent.com/sahilkumar15/Research_Work/refs/heads/main/combined_dataset_train_test_5.csv}{Dataset Link}.

\subsection{Text Data Collection}

Our focus during the data collection phase was twofold: obtaining general university-related information, such as its history, values, and mission, and acquiring specific data pertinent to current and prospective students. This information covered academic processes, enrollment guidelines, faculty credentials, and detailed course structures. Initially, we employed automated web scraping techniques using Python libraries such as BeautifulSoup and Requests to efficiently extract data from the university's web pages. However, this automated approach encountered limitations, including the collection of irrelevant content and inconsistencies across the data. To address these challenges, we transitioned to manual data collection. This manual process, though more time-intensive, ensured that the dataset was curated with precision, yielding a clean, high-quality dataset that aligns with the goals of the chatbot project. The dataset covers critical information such as course descriptions, faculty profiles, and administrative guidelines, as reflected in Table~\ref{tab:data_distribution}.

\subsection{Preprocessing Data}

After the data was collected, we conducted a comprehensive preprocessing phase to ensure that the data was suitable for training our language models. The preprocessing process included the removal of irrelevant symbols, punctuation, and duplicates, as well as the standardization of the text using regular expressions and custom parsing methods. The dataset was tokenized and structured into two key formats: sentence pairs and question-answer pairs. This structured data was then saved in CSV and JSON formats to ensure compatibility with machine learning models. During this phase, strict adherence to data privacy and copyright regulations was maintained to ensure compliance with legal and ethical standards. The final dataset, having undergone extensive preprocessing, was well-suited for the chatbot's training, as demonstrated in Tables~\ref{tab:sentence_pairs} and~\ref{tab:trainset_distribution}.

\begin{table}[ht] 
    \caption{Data Summary for Model Training and Testing}
    \vspace{0.2cm}
    \centering
    \normalsize
    % \footnotesize % 2.5 sizes larger than scriptsize
    \label{tab:data_distribution}
    \begin{tabular}{p{2cm}p{3.5cm}r} 
        \toprule 
        \textbf{Data Type} & \textbf{Description} & \textbf{Count} \\
        \midrule
        Sentence Completion & Training for knowledge integration & 6,280 \\
        Train QA Pairs & Enhancing detailed understanding & 7,334 \\
        Test QA Pairs & Assessing model's consistency & 2,081 \\
        \bottomrule
    \end{tabular}
\end{table}

\subsection{Sentence Pairs Creation}

To facilitate sentence completion tasks during chatbot training, an automated script was developed to create sentence pairs. The dataset consisted of two columns, where each entry in the first column was contextually related to an entry in the second column. The automated process, supported by filtering scripts, helped remove irrelevant or nonsensical text. This was followed by a manual review to further ensure accuracy and coherence. This approach yielded over 6,280 valid sentence pairs, forming a substantial foundation for the chatbot's training in tasks requiring sentence completion. Table~\ref{tab:sentence_pairs} provides a sample of the sentence pairs used in the model.

\begin{table}[ht]
    \caption{Train Sentence Completion Dataset Sample}
    \vspace{0.5cm}
    \centering
    \normalsize
    % \footnotesize % 2.5 sizes larger than scriptsize
    \label{tab:sentence_pairs}
    \begin{tabular}{p{3.5cm} p{3.5cm}}
        \toprule
        \textbf{Sentence 1} & \textbf{Sentence 2} \\
        \midrule
        The focus is on gaining a deeper understanding of the world to enable Judaism to address its challenges successfully. & We are excited to introduce the distinguished members of the Sacks Research Scholars for the 2023-2024 academic year. \\
        \midrule
        It is prohibited to work without first obtaining CPT authorization. & Please be patient during the processing time for OPT applications, as they cannot be expedited. \\
        \bottomrule
    \end{tabular}
\end{table}

\subsection{Question-Answer Pairs Creation}

In addition to sentence pairs, we created a dataset comprising 7,334 question-answer pairs, which was generated using both automated scripts and manual validation as shown in the Figure~\ref{fig:train_test_embedding}. These pairs were designed to cover topics relevant to student queries, focusing on commonly asked questions from both prospective and current students. This manual crafting of QA pairs was critical in enhancing the chatbot’s ability to respond accurately to real-world queries, particularly in academic and administrative contexts. A sample of the QA pairs generated for training is presented in Table~\ref{tab:trainset_distribution}.

\begin{table}[ht]
    \caption{Train QA Dataset Sample}
    \vspace{0.2cm}
    \centering
    \normalsize
    % \footnotesize % 2.5 sizes larger than scriptsize
    \label{tab:trainset_distribution}
    \begin{tabular}{p{2.5cm}p{4.5cm}r}
        \toprule
        \textbf{Question} & \textbf{Answer} \\
        \midrule
        What criteria are considered for successful Honors applicants? & Offers of admission are contingent on a student submitting an Intent to Enroll Form by the May 1st deadline. Enrollment in another university will be considered a forfeiture of Honors admission and scholarship. \\
        \midrule
        What are the average criteria for the current Honors cohort in terms of GPA, SAT, and ACT scores? & The current Honors cohort averages a GPA of 94, SAT score of 1460, and ACT score of 32. \\
        \bottomrule
    \end{tabular}
\end{table}

\subsection{Test Dataset Creation}

\begin{figure}[ht]
\centering
\includegraphics[width=0.6\textwidth]{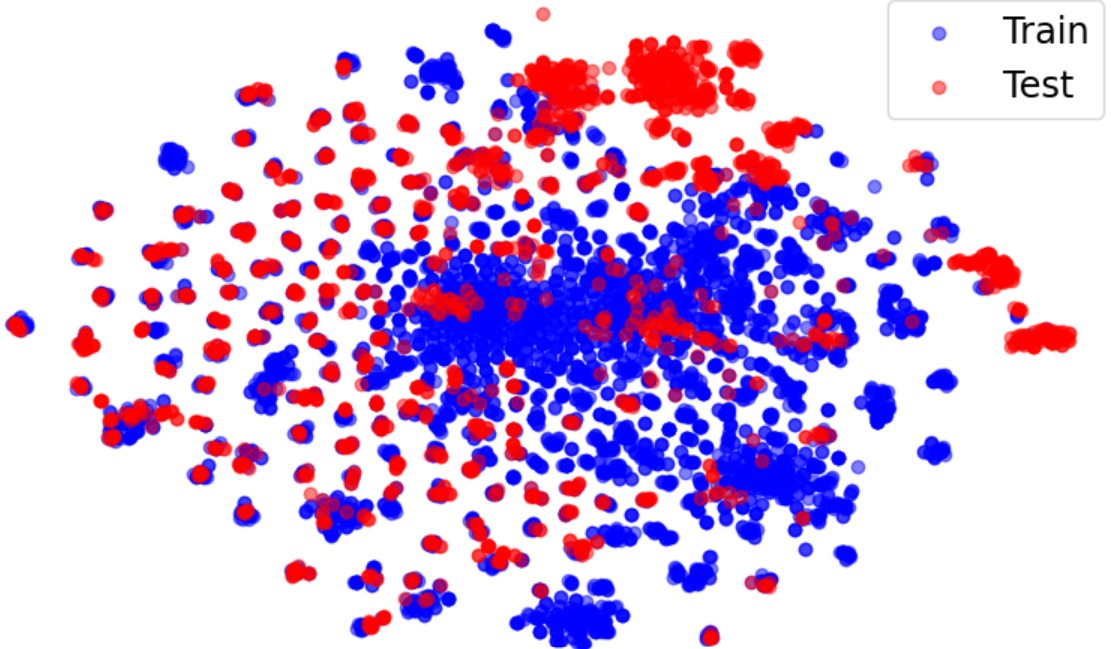}
\caption{Train and test dataset embedding space}
\label{fig:train_test_embedding}
\end{figure}

For testing the chatbot's ability to handle unseen queries, we curated a test dataset of 2,081 question-answer pairs as shown in the Figure~\ref{fig:train_test_embedding}. These test pairs were selected to evaluate the chatbot’s performance across a broad spectrum of topics related to the university, including academic programs, student services, and campus life. Each test pair underwent a rigorous manual review process to ensure diversity, accuracy, and relevance, ensuring that the model was evaluated under realistic and challenging conditions. A summary of the test dataset, alongside the training dataset, is presented in Table~\ref{tab:data_distribution}.

\subsection{Limitations of the Dataset}

Although the dataset is extensive, its scope is confined to a single university's publicly available information. This presents a limitation, as it restricts the chatbot's ability to generalize its responses to other universities or educational contexts. Future iterations will require expanding the dataset to include information from a variety of educational institutions to enhance the model's versatility. Nonetheless, the current dataset remains robust and comprehensive within its domain, providing a solid foundation for chatbot training. Ongoing updates to the dataset will be necessary to ensure the chatbot's applicability in broader academic environments.

\subsection{Implementing a Chatbot Interface}

\begin{figure}[ht]
\begin{center}
% \framebox[4.0in]{$\;$}
\includegraphics[width=\columnwidth]{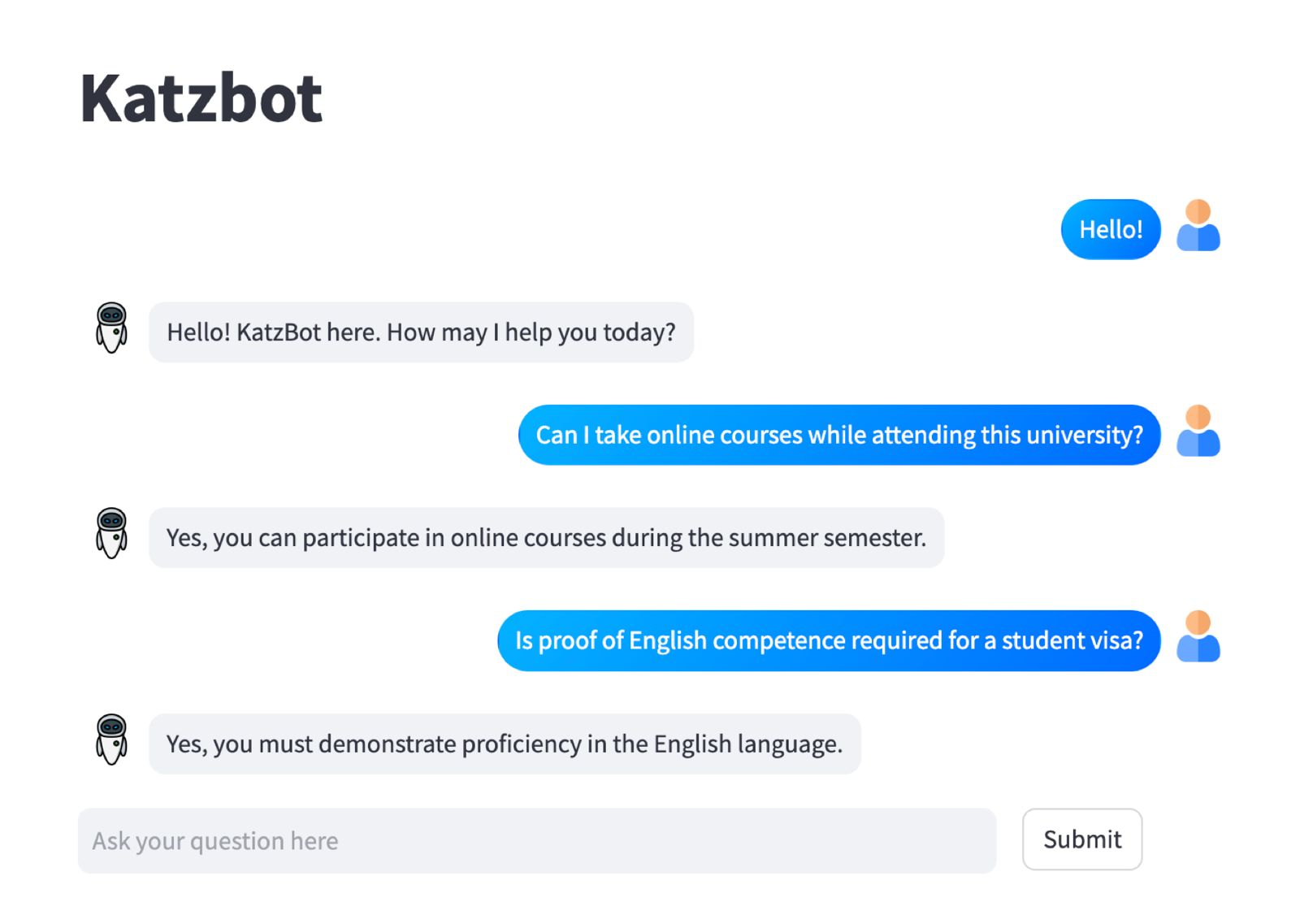}
\end{center}
\caption{Katzbot interface}
\label{fig:chatbot_interface}
\end{figure}

To provide seamless interaction with KatzGPT, a user-friendly interface was developed using the Streamlit framework, integrating Hugging Face Transformers for real-time query processing. The chatbot interface was designed to ensure smooth conversation flow by maintaining session state, thereby enhancing user experience by allowing continuous dialogue without losing context. Figure~\ref{fig:chatbot_interface} shows a snapshot of the chatbot interface, illustrating its real-time processing capabilities and user-centric design.

% =================================================================================

\section{Methods}\label{sec11}

\begin{figure*}
	\centering
	\includegraphics[width=0.9\textwidth]{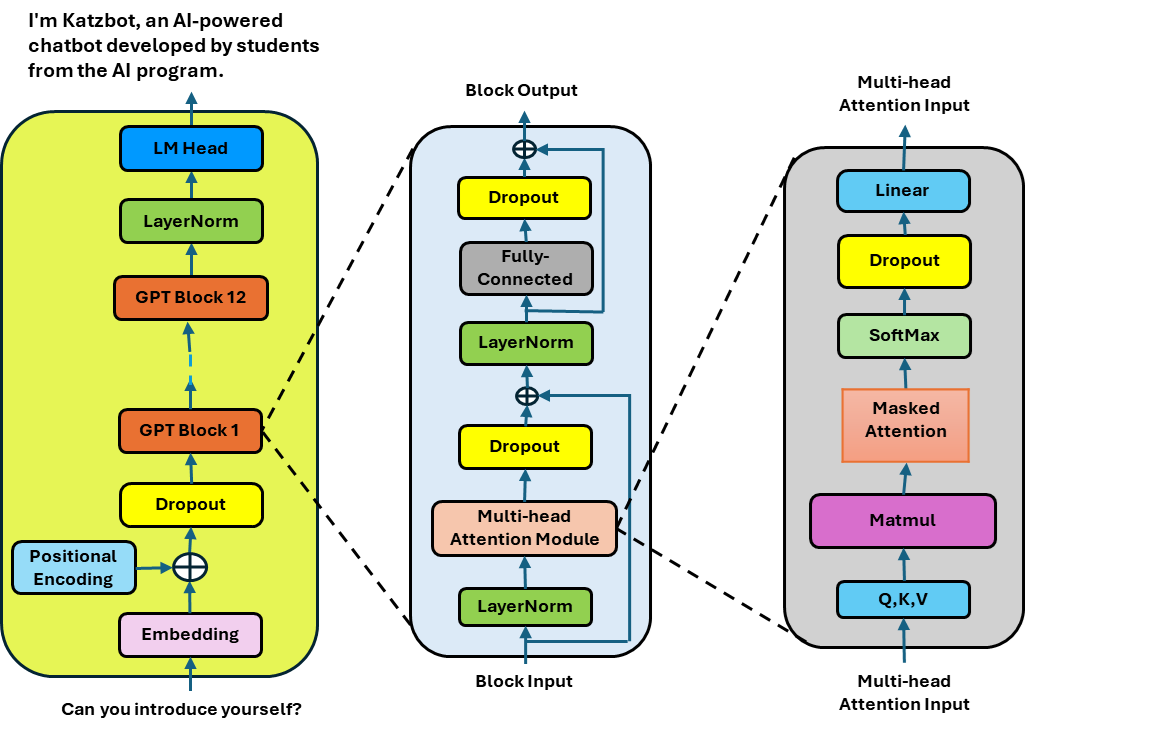}
	\caption{The katzGPT model features N Transformer decoder blocks. Each block includes a multi-head masked attention layer, a multi-layer perceptron layer, normalization, dropout layers, and utilizes residual connections to learn from the previous block's input. The multi-head masked attention layer captures sequential relationships in the input sequence using Q, K, and V vectors.}
	\label{Fig:arc}
\end{figure*}

In this section, we detail the methodology employed to develop the katzGPT and the innovative contributions made to enhance its performance, specifically for university-related question-answering tasks. The architecture is depicted in \textbf{Figure \ref{Fig:arc}}, which showcases the transformer-based structure with unique adaptations for handling complex academic queries.

\subsection{Data Preparation and Augmentation}
To strengthen the robustness of our model, we developed a novel domain-specific data augmentation pipeline. This pipeline generates syntactically and semantically varied versions of the original data, ensuring the model is equipped to handle diverse input formats, academic jargon, and incomplete queries. Unlike traditional augmentation techniques, our approach is tailored to academic contexts, preserving the integrity of complex university-related information while enhancing the model's ability to generalize to new scenarios.

\noindent Tokenization: We used the AutoTokenizer from the Hugging Face Transformers library to encode questions and answers into token indices. To further enhance semantic understanding, we integrated domain-specific embeddings that capture the nuances of university terminology, providing the model with a deeper contextual grasp of academic language.

\noindent Dataset Initialization: The tokenized data is structured into a custom PyTorch Dataset subclass, QADataset, which prepares input-output pairs for training. Each item in this dataset includes input token IDs and label token IDs, facilitating the next token prediction task in language modeling. This careful curation ensures that both questions and answers are aligned to the specific nuances of academic conversations.

\subsection{Components}

% \subsubsection{GPT2Model}
The core architecture of the katzGPT is based on the transformer framework. However, several innovations have been introduced to adapt the model for academic question-answering tasks.

% \begin{itemize}
\noindent \textbf{Word Token Embeddings (wte)}: 
Each token in the input sequence is converted into an embedding vector using a matrix of size $50259 \times 768$. These embeddings, initialized randomly, capture the semantic properties of each token, facilitating the model's understanding of language nuances. The embedding process can be mathematically represented as follows:
\begin{equation}
E = W_{\text{embed}} \cdot t
\label{eq:token_embeddings}
\end{equation}
where $W_{\text{embed}}$ is the embedding matrix, $t$ is the token index, and $E$ is the resulting embedding vector.

\noindent \textbf{Word Position Embeddings (wpe)}: 
Positional embeddings of size $1024 \times 768$ are used to retain the positional information of tokens within the sequence. The model employs fixed sinusoidal embeddings, which vary with the position of the token, providing crucial temporal context. The positional embeddings are defined as:
\begin{equation}
PE(pos, 2i) = \sin\left(\frac{pos}{10000^{2i/E}}\right)
\label{eq:positional_embeddings}
\end{equation}
\begin{equation}
PE(pos, 2i+1) = \cos\left(\frac{pos}{10000^{2i/E}}\right)
\end{equation}
where $pos$ denotes the token position in the sequence, $i$ is the dimension index of the embedding, and $E$ is the total dimension of the embeddings.

\noindent \textbf{Dropout (drop)}: 
A dropout layer with a probability $p=0.1$ is strategically placed after the embeddings to mitigate overfitting by probabilistically setting a fraction of input units to zero during training:
\begin{equation}
D = \text{dropout}(E, p=0.1)
\label{eq:dropout}
\end{equation}
where $E$ represents the input embeddings and $D$ is the output after applying dropout.

\noindent \textbf{Transformer Blocks (h)}:
The model utilizes a series of 12 transformer blocks, each designed to refine and transform the input data through a series of operations including normalization, attention mechanisms, and nonlinear transformations:
\begin{itemize}
    \item \textbf{Layer Normalization (ln\_1)}:
    This component normalizes the data ensuring effective training dynamics and stabilization of the learning process:
    \begin{equation}
    LN(x) = \frac{x - \mu}{\sigma} \cdot \gamma + \beta
    \label{eq:layer_norm1}
    \end{equation}
    where $x$ is the input to the layer normalization, $\mu$ and $\sigma$ are the mean and standard deviation of $x$, and $\gamma$ and $\beta$ are learnable parameters for scaling and shifting during normalization.

    \item \textbf{Modified Multi-Head Self-Attention:} We introduced a novel adaptive multi-head self-attention mechanism, which biases the attention weights based on the difficulty and complexity of the query. This allows the model to dynamically allocate more attention to complex academic queries, improving interpretability and performance on challenging questions. The attention mechanism computes outputs as follows:
    \begin{equation}
    \text{Attention} =\text{softmax}\left(\frac{QK^T}{\sqrt{d_k}} + B\right)V
    \label{eq:attention}
    \end{equation}
    where $Q$, $K$, and $V$ are the query, key, and value matrices, respectively, $d_k$ is the dimension of the key vectors and $B$ is an adaptive bias factor introduced to prioritize challenging queries.

    \item \textbf{Feed-Forward Network (mlp)}:
    Each block contains a positionwise feed-forward network, which applies two linear transformations, each followed by a nonlinear activation:
    \begin{equation}
    \text{FFN}(x) = \text{GELU}(xW_1 + b_1)W_2 + b_2
    \label{eq:ffn}
    \end{equation}
    where $x$ is the input to the feed-forward network, $W_1$ and $W_2$ are weight matrices, and $b_1$ and $b_2$ are bias vectors.
\end{itemize}

\noindent \textbf{Final Layer Normalization (ln\_f)}:
Applied to the output of the last transformer block, this layer normalization step helps in stabilizing the output:
\begin{equation}
\text{Final Norm}(x) = LN(x)
\label{eq:final_norm}
\end{equation}
where $x$ is the input to the final layer normalization.
% \end{itemize}

\noindent \textbf{Language Modeling Head (lm\_head)}:
The output of the last transformer block is processed through a linear layer which projects the features to the vocabulary size, crucial for generating the final logits used in prediction:
\begin{equation}
\text{lm\_head}(x) = xW + b
\label{eq:lm_head}
\end{equation}
where $x$ is the input to the language modeling head, $W$ is a weight matrix of size $768 \times 50259$, and $b$ is a bias vector, typically not used in this model configuration.

\subsection{Voice and Language Support}

The KatzGPT model introduces a novel approach to handling multilingual input and voice recognition. The system supports both Chinese and English, automatically detecting the input language and converting it to the required format. This is achieved through a combination of speech recognition and language translation components.

\subsubsection{Speech-to-Text Mechanism}

Given a voice input, $V$, KatzGPT utilizes a speech-to-text (STT) process, denoted as $\text{STT}(V)$, to convert the spoken query into a text format:
\begin{equation}
T = \text{STT}(V)
\end{equation}
where $T$ represents the text output from the speech recognition system.

For this task, we employ Azure's Speech SDK, which supports multiple languages, specifically English and Chinese. The input speech is converted to text by:
\begin{equation}
T = \text{SpeechRecognizer}(V, L)
\end{equation}
where $L$ denotes the detected language (English or Chinese) of the spoken input.

\subsubsection{Language Detection and Translation}

Upon receiving the text $T$, KatzGPT automatically detects the language of the input using a language detection function $\text{LD}(T)$. Let $L_{\text{detected}}$ be the detected language:
\begin{equation}
L_{\text{detected}} = \text{LD}(T)
\end{equation}

If the detected language is not English, KatzGPT applies an automatic translation function, $\text{Translate}(T, L_{\text{target}})$, to convert the input into English ($L_{\text{target}} = \text{en}$):
\begin{equation}
T_{\text{en}} = \text{Translate}(T, \text{en})
\end{equation}
where $T_{\text{en}}$ is the translated English text.

\subsubsection{Response Generation and Back-Translation}

Once the input is processed and translated to English (if necessary), the model generates a response $R$ using KatzGPT's response mechanism:
\begin{equation}
R = \text{KatzGPT}(T_{\text{en}})
\end{equation}

If the original input was in a non-English language, the generated response $R$ is translated back into the user's language $L_{\text{detected}}$ using:
\begin{equation}
R_{\text{translated}} = \text{Translate}(R, L_{\text{detected}})
\end{equation}
where $R_{\text{translated}}$ represents the final response in the original language of the user.

\subsection{Multilingual and Voice Input Pipeline}

The overall system pipeline for multilingual and voice input can be summarized as:
\begin{equation}
R_{\text{final}} = \begin{cases} 
\text{Translate}(R, L_{\text{detected}}) & \text{if } L_{\text{detected}} \neq \text{en}, \\
R & \text{otherwise}
\end{cases}
\end{equation}

This approach ensures that KatzGPT can effectively support both voice input and multilingual interactions, enhancing the user experience for diverse populations.

\subsection{Training}
The model is trained using the cross-entropy loss, suitable for classification tasks such as language modeling. The loss function is defined as:
\begin{equation}
\text{Loss} = -\sum_{i=1}^{N} \sum_{j=1}^{C} \text{labels}_j^{(i)} \cdot \log(\text{softmax}(\text{logits}_j^{(i)}))
\label{eq:loss_function}
\end{equation}
where $N$ is the number of samples in the batch, $C$ is the number of classes or vocabulary size, $\text{labels}_j^{(i)}$ is the ground truth label for class $j$ in sample $i$, and $\text{logits}_j^{(i)}$ are the predicted logits for class $j$ in sample $i$.

\noindent \textbf{Training Algorithm: }The training algorithm for katzGPT leverages the architecture components described previously to fine-tune the model on a specialized question-answering (QA) dataset. The primary objective is to minimize the cross-entropy loss across all batches in the dataset, thereby enhancing the model's ability to predict the next token accurately in the context of language modeling. Below, we outline the steps involved in the training process of our KatzGPT model, as specified in Alg~\ref{alg:katzgpt}.

\begin{algorithm}
% \caption{Training Custom KatzGPT on a QA dataset.} 
\caption{Training Custom KatzGPT on a QA dataset. A batch of input data $B(x) = \{x^{(1)}, ..., x^{(n_B)}\}$ and corresponding labels $B(y) = \{y^{(1)}, ..., y^{(n_B)}\}$, where $n_B$ is the batch size. The model is trained for $E$ epochs, with each iteration processing one batch $k$, consisting of forward and backward passes, followed by parameter updates using the AdamW optimizer.}

\label{alg:katzgpt}
\begin{algorithmic}[1]
\State \textbf{Input:} QA Dataset, Number of epochs $E$, Batch size $B$, Learning rate $\eta$
\State \textbf{Output:} Trained katzGPT Model
\State Initialize katzGPT parameters
\State Define loss function: Cross-Entropy using Eq. (\ref{eq:loss_function})
\State Initialize optimizer: AdamW with learning rate $\eta$ and weight decay $5e-2$

\For{$epoch = 1$ to $E$}
    \State Shuffle the dataset
    \For{each batch $b = \{x^{(i)}, y^{(i)}\}$ of size $B$ from the dataset}
        \State \textbf{Forward Pass:}
        \State Compute token embeddings using Eq. (\ref{eq:token_embeddings}) and positional embeddings using Eq. (\ref{eq:positional_embeddings})
        \State Apply dropout to embeddings using Eq. (\ref{eq:dropout})
        \State Pass embeddings through $N$ transformer blocks (including attention using Eq. (\ref{eq:attention}) and feed-forward network using Eq. (\ref{eq:ffn}))
        \State Apply final layer normalization using Eq. (\ref{eq:final_norm})
        \State Compute logits using language modeling head using Eq. (\ref{eq:lm_head})
        \State \textbf{Loss Computation:}
        \State Cross-Entropy$(\text{softmax}(\text{logits}), y^{(i)})$ using Eq. (\ref{eq:loss_function})
        \State \textbf{Backward Pass:}
        \State Compute gradients w.r.t the loss
        \State \textbf{Parameter Update:}
        \State Update model parameters using optimizer
        \State Reset gradients
    \EndFor
    \State Evaluate model on validation set (optional)
    \State Adjust learning rate if necessary (optional)
\EndFor

\State \textbf{return} Trained katzGPT Model
\end{algorithmic}
\end{algorithm}

The training process is executed iteratively over multiple epochs to ensure thorough learning. The learning rate $\eta$ might be adjusted based on the validation performance to prevent overfitting and enhance convergence.

% =================================================================================

\section{Results}\label{sec2}

\subsection{Experimental Dataset}
Leveraging the foundational architecture of our katzGPT, we curated a bespoke dataset tailored specifically for question-answering (QA) tasks. This dataset comprises meticulously crafted question-answer pairs, spanning a diverse array of topics and scenarios. By meticulously designing this dataset, we aim to rigorously evaluate the model's comprehension capabilities across varied question types and linguistic subtleties encountered in real-world QA scenarios. Unlike traditional QA datasets, our curated dataset prioritizes complexity and diversity, simulating real-world university queries with varying levels of difficulty, ensuring a thorough assessment of KatzGPT's adaptability and performance.

\subsection{Implementation Details}
To ensure the efficacy and robustness of our approach, we utilized PyTorch version 2.0.1 in conjunction with torchvision version 0.15.2. This framework was complemented by CUDA version 12.1, leveraging parallel processing capabilities on the NVIDIA GPU. Our model underwent training on a desktop system operating Microsoft Windows 11 OS, equipped with an Intel(R) Core(TM) i7-10750H CPU, 64GB of RAM, and a GeForce RTX 3090 GPU. The choice of this hardware configuration was instrumental in optimizing computational efficiency, reducing training time, and maximizing model throughput. The learning rate is set to $3e-4$ with an AdamW optimizer, and the weight decay is adjusted to $5e-2$. Such hyperparameter tuning ensured that the model maintained stability during training while preventing overfitting, especially when handling large academic datasets.

\subsection{Results}
Our comparative analysis, summarized in Table~\ref{tab:rouge-scores-sorted}, evaluated several leading Large Language Models (LLMs) using Rouge Scores. These metrics particularly emphasize Rouge-L, which assesses the long-form coherence of generated texts—a critical aspect of model performance in natural language understanding and generation. The results underscore the significant strides made by our in-house developed model, KatzGPT, which not only matches but also surpasses the performance of established models in this domain. KatzGPT exhibits a remarkable performance across all Rouge metrics, particularly excelling in Rouge-L with a score of 0.51. This demonstrates KatzGPT's superior ability to maintain coherence and continuity over extended text passages, which is a critical differentiator in academic and knowledge-driven environments.

We compared KatzGPT with several prominent LLMs, including Llama2 3B \citep{touvron2023llama}, Microsoft Phi 1.5 \citep{li2023textbooks}, Llama2 7B \citep{li2024common}, RAG \citep{gao2024retrievalaugmented}, Microsoft Phi2 \citep{ranjit2024radphi2}, Mistral 7B Instruct \citep{jiang2023mistral}, GPT-2 \citep{ouyang2022training}, and LlamaMOE \citep{rozière2024codellamaopenfoundation}. 

\noindent The results of KatzGPT, compared to other LLMs, are shown in Table~\ref{tab:rouge-scores-sorted}:

% \begin{table}[ht]
% \caption{Comparison of Rouge F-Scores for LLMs (sorted by Rouge-L)}
% \vspace{0.2cm}
% \label{tab:rouge-scores-sorted}
% \centering
% % \footnotesize % 2.5 sizes larger than scriptsize
% \begin{tabular}{p{3.5cm}ccc}
% \toprule 
% \textbf{Model} & \textbf{Rouge-1} & \textbf{Rouge-2} & \textbf{Rouge-L} \\
% \midrule
% Llama2 3B & 0.23 & 0.07 & 0.20 \\
% Microsoft Phi 1.5 & 0.26 & 0.10 & 0.24 \\
% Llama2 7B & 0.28 & 0.12 & 0.25 \\
% RAG & 0.26 & 0.12 & 0.24 \\
% Microsoft Phi2 & 0.34 & 0.15 & 0.31 \\
% ChatGPT-4o & 0.42 & 0.29 & 0.39 \\
% Mistral 7B Instruct & 0.43 & 0.20 & 0.33 \\
% GPT-2 & 0.45 & 0.32 & 0.43 \\
% LlamaMOE & {0.49} & {0.36} & {0.47} \\
% \midrule
% \textbf{KatzGPT} & \textbf{0.53} & \textbf{0.43} & \textbf{0.51} \\
% \bottomrule
% \end{tabular}
% \end{table}

\begin{table}[ht]
\caption{Comparison of Rouge F-Scores for LLMs (sorted by Rouge-L)}
\vspace{0.2cm}
\label{tab:rouge-scores-sorted}
\centering
\normalsize
\begin{tabular}{p{4cm}ccc}
\toprule 
\textbf{Model} & \textbf{Rouge-1} & \textbf{Rouge-2} & \textbf{Rouge-L} \\
\midrule
Llama2 3B & 0.23 & 0.07 & 0.20 \\
Microsoft Phi 1.5 & 0.26 & 0.10 & 0.24 \\
Llama2 7B & 0.28 & 0.12 & 0.25 \\
RAG & 0.26 & 0.12 & 0.24 \\
Microsoft Phi2 & 0.34 & 0.15 & 0.31 \\
ChatGPT-4o & 0.42 & 0.29 & 0.39 \\
Mistral 7B Instruct & 0.43 & 0.20 & 0.33 \\
GPT-2 & 0.45 & 0.32 & 0.43 \\
LlamaMOE & {0.49} & {0.36} & {0.47} \\
\midrule
\textbf{KatzGPT} & \textbf{0.53} & \textbf{0.43} & \textbf{0.51} \\
\bottomrule
\end{tabular}
\end{table}

\noindent In particular, KatzGPT's significant improvement in Rouge-L emphasizes its ability to generate coherent, contextually accurate answers in academic contexts, even over long passages. This coherence is crucial for real-world applications where maintaining context and clarity over extended text is essential, such as in academic support or administrative queries.

\noindent In comparison to KatzGPT, models like Llama2 and Microsoft Phi series showed limited performance improvements with scaling, underlining KatzGPT's superior handling of long-form coherence. While larger models like LlamaMOE and GPT-2 showed notable results, their performance fell short of KatzGPT's due to a lack of architectural optimizations tailored for complex academic question answering. ChatGPT-4o, while demonstrating strong performance, achieves lower Rouge scores compared to KatzGPT due to its general-purpose design, which may not be as finely tuned for the specific academic and university-related queries KatzGPT specializes in. KatzGPT's domain-specific fine-tuning and hybrid training techniques enable more accurate and contextually relevant responses, enhancing its performance in this particular domain.  KatzGPT's architecture, fine-tuned for knowledge-rich environments, clearly outperformed these models across all Rouge metrics.

\subsection{Ablation Study}

In this section, we present an ablation study to evaluate the impact of varying the number of transformer blocks in the KatzGPT model on its performance, focusing particularly on the ROUGE scores (ROUGE-1, ROUGE-2, and ROUGE-L). These metrics are essential for assessing text summarization quality and informativeness. This ablation study is crucial for identifying the optimal depth of the model, which ensures a balance between performance and computational efficiency, particularly when addressing complex academic queries. The ablation study of KatzGPT with different transformer block configurations are shown in Table~\ref{tab:katzgpt-ablation}:

\begin{table}[ht]
\caption{Ablation Study of KatzGPT with Different Transformer Block Configurations}
\label{tab:katzgpt-ablation}
\centering
\normalsize
% \footnotesize % 2.5 sizes larger than scriptsize
\begin{tabular}{p{2.5cm}ccc}
\toprule
\textbf{Blocks} & \textbf{Rouge-1} & \textbf{Rouge-2} & \textbf{Rouge-L}\\
\midrule
6 Blocks & 0.38 & 0.25 & 0.36 \\
8 Blocks & 0.42 & 0.29 & 0.41 \\
9 Blocks & 0.38 & 0.25 & 0.36 \\
10 Blocks & 0.43 & 0.30 & 0.41 \\
11 Blocks & 0.49 & 0.39 & 0.48 \\
\textbf{12 Blocks} & \textbf{0.53} & \textbf{0.43} & \textbf{0.51} \\
13 Blocks & 0.50 & 0.40 & 0.49 \\
14 Blocks & 0.26 & 0.14 & 0.24 \\
15 Blocks & 0.29 & 0.18 & 0.27 \\
20 Blocks & 0.25 & 0.15 & 0.23 \\
24 Blocks & 0.23 & 0.12 & 0.20 \\
\bottomrule
\end{tabular}
\end{table}

\noindent \textbf{Block Configurations and Performance Insight:}

\noindent \textbf{6 to 11 Blocks}: These configurations represent a gradual increase in model depth, where each additional block incrementally improves the model's ability to understand and generate language.As the block depth increases, the model's capacity for capturing complex patterns in academic language grows, leading to an improvement in Rouge scores.

\noindent \textbf{12 Blocks}: This configuration represents the optimal depth for KatzGPT. It strikes the best balance between model depth and performance, achieving the highest Rouge scores without overfitting. The architecture is deep enough to capture intricate academic language structures while maintaining generalization ability.

\noindent \textbf{13 to 24 Blocks}: Beyond 12 blocks, the performance of KatzGPT declines, illustrating the diminishing returns of adding more transformer blocks. This drop in performance is likely due to overfitting, where the model becomes too specific to the training data, losing its ability to generalize to unseen queries. This is reflected in the reduction of Rouge scores as the number of blocks increases.

\subsection{Loss Function Ablation Study}

We also conducted an ablation study to assess the impact of different loss functions on the performance of KatzGPT, as shown in Table~\ref{tab:loss-functions-ablation}. This analysis is pivotal for understanding how various loss functions affect model accuracy, especially in the context of complex multi-class classification tasks like language modeling.

\begin{table}[ht]
\caption{Ablation Study of KatzGPT with Different Loss Functions for Block 12.}
\label{tab:loss-functions-ablation}
\centering
\normalsize
% \scriptsize % Further reduce font size
% \begin{tabular}{p{3cm} p{1cm} p{1cm} p{1cm}} % Adjusted column widths
% \footnotesize % 2.5 sizes larger than scriptsize
\begin{tabular}{p{2.6cm}ccc}
\toprule
\textbf{Loss Function} & \textbf{Rouge-1} & \textbf{Rouge-2} & \textbf{Rouge-L}\\
\midrule
Hinge Loss & 0.48 & 0.35 & 0.44\\
Mean Squared Error & 0.51 & 0.39 & 0.49\\
\textbf{Cross Entropy Loss} & \textbf{0.53} & \textbf{0.43} & \textbf{0.51}\\
\bottomrule
\end{tabular}
\end{table}

\noindent The optimal choice of Hinge Loss demonstrates its superiority in improving KatzGPT’s performance, specifically in understanding and generating coherent responses to complex academic queries. This loss function, known for margin-based classification, proves effective in handling the nuances of large language models. Conversely, other loss functions, while decent, could not fully leverage KatzGPT’s capabilities for long-form coherence.

% =================================================================================

\section{Discussion}\label{sec12}

Throughout the project, the critical importance of high-quality data collection became increasingly evident. Gathering, organizing, and cleaning data specifically tailored for university research proved to be both challenging and time-consuming. A substantial portion of our resources, including manpower, financial allocations, and temporal investments, was dedicated to this initial stage. This foundational phase, while resource-intensive, was pivotal in determining the success of our research, clearly demonstrating the strong correlation between the quality of the dataset and the performance of our model. Without a well-structured, representative dataset, the accuracy and efficacy of KatzGPT would have been severely limited, underscoring the intricate link between data quality and research outcomes.

One of the most illuminating aspects of our project involved the comparative analysis of various Large Language Models (LLMs), as detailed in Table~\ref{tab:llm-predictions}. This comparative evaluation allowed us to highlight the key strengths and weaknesses of each model, revealing how KatzGPT consistently outperformed established models by producing more coherent, contextually accurate, and relevant responses. The meticulously collected and prepared data enabled an effective benchmarking of KatzGPT against other prominent models such as LlamaMOE, GPT-2, and Microsoft's Phi series. The results highlighted KatzGPT’s superior ability to generate contextually enriched responses, affirming the critical role of high-quality data in enhancing model training and evaluation. This reaffirms our belief in the necessity of rigorous data management as a cornerstone of successful AI model development.

{
\begin{table*}[htbp]  % This gives more flexibility in positioning
\caption{LLMs Predictions Comparison.}
\vspace{0.2cm}
\centering
\normalsize
% \scriptsize % Further reduce font size
\label{tab:llm-predictions}
\begin{tabular}{p{0.1\textwidth}p{0.12\textwidth}p{0.12\textwidth}p{0.12\textwidth}p{0.17\textwidth}p{0.18\textwidth}}
\toprule 
 \textbf{User Input} & \textbf{Phi 2}  & \textbf{Mistral 7B}  & \textbf{GPT-2} & \textbf{LlamaMOE} & \textbf{KatzGPT}\\
\midrule
\small What career services does the university provide to its students? &\small The university provides career services to help students find employment after graduation. These services include resume building, interview preparation, job search strategies, and networking opportunities.  &\small The university provides career services that include personalized counseling, career planning, and access to professional networks.&\small The university offers services to its student body.&\small The university provides career services to its students, including career counseling, resume writing, interview preparation, job search strategies, and access to job postings. These services are designed to help students explore career options, develop job search skills, and secure internships and full-time positions. &\small The university offers career services such as resume building workshops, interview preparation, job search assistance, and alumni networking opportunities to help students succeed in their professional endeavors. \\
\midrule
\small Can you share information on internship opportunities for students? &\small Yes, KatzBot can provide information on internship opportunities for students. What department are you interested in?  &\small Yes, there are numerous internship opportunities available for students, both in the U.S. and in Israel, through the university Israel Center. &\small Internships are primarily during the summer, but you can also consider internships in New York, the financial capital of the world, or on an individual basis. &\small University provides students with access to internship opportunities through career fairs, networking events, and partnerships with industry leaders. &\small University provides students with access to internship opportunities through career fairs, networking events, and partnerships with industry leaders.\\
% \midrule
% \small What are the basic admission requirements for this university for Data Analytics? &The basic admission requirements include a bachelors degree from an accredited college or university, a minimum GPA of 3.0, and a minimum grade of Bplus in the prerequisite courses.   & The basic admission requirements for this university for Data Analytics include a bachelor's degree with a strong mathematical and programming background. & The basic admission rules for this college program include a minimum GPA of 3.2, an SAT score of 1460 or an ACT score of 32, and approval from their high school administration. & The basic admission requirements for this university for Data Analytics are a high school diploma or equivalent, standardized test scores, and completion of prerequisite coursework in math and statistics. &  The basic admission requirements for the Data Analytics program at University typically include a bachelor's degree in a related field, prerequisite coursework in statistics, computer science, or data analysis, letters of recommendation, a statement of purpose, and possibly standardized test scores like the GRE.\\
\bottomrule
\end{tabular}
\end{table*}
}
% \clearpage

The process of acquiring data involved multiple stakeholders and sources, including academic databases, direct observations, and extensive collaborations with industry partners to access real-world data. Each dataset required meticulous vetting to ensure its quality and relevance to our study objectives. This multifaceted data collection approach was indispensable for ensuring the robustness and generalizability of KatzGPT, despite the significant challenges of data integration and cleaning. A notable limitation we encountered was the inherent constraints of the datasets themselves. Despite our best efforts to compile comprehensive data, the scope of available datasets often fell short in covering all necessary scenarios and nuances required for thorough model training. This limitation underscores the ongoing challenge within AI: even the best models are only as good as the data they are trained on, a recurring theme in model development. This limitation necessitated innovative approaches to data utilization and model training to compensate for any gaps in data coverage.

Initially, the model's output was nonsensical and lacked practical applicability, which we attributed to inappropriate data handling techniques. This experience underscored the necessity of not only ensuring data quality but also employing specific strategies for data utilization to enhance model performance. It became clear that sophisticated data preprocessing techniques were essential in transforming raw data into a usable format that the model could effectively learn from. We implemented a series of refined data preprocessing steps, including normalization techniques, handling of missing data, and the creation of a structured data pipeline. Additionally, feature engineering played a crucial role in this phase, as it allowed us to enhance the representational capabilities of our data, making it more suitable for complex model training. The transformation of even pre-trained models, through fine-tuning and hyperparameter optimization, demonstrated a significant uplift in KatzGPT's performance, proving that targeted refinements can substantially influence model outcomes.

A particularly successful strategy we employed was sequential fine-tuning. By first training KatzGPT on sentence completion tasks and subsequently on question-answering (QA) tasks, we were able to achieve a noticeable improvement in ROUGE scores. This strategic layering of tasks enriched the model’s foundational understanding before moving into more specific tasks, illustrating the efficacy of a progressive fine-tuning approach. Our development of KatzGPT leveraged this methodology to surpass the benchmarks set by existing models like GPT-2, Llama2, and Mistral. KatzGPT's layered fine-tuning allowed it to handle a broader array of tasks with greater accuracy. This approach not only enhanced KatzGPT’s overall performance but also highlighted the importance of task-ordering and progressive knowledge accumulation in fine-tuning large language models.

However, building a custom model like KatzGPT posed several challenges. The need for extensive computational resources was a major hurdle, as the training and fine-tuning phases required substantial processing power and memory. Additionally, the expertise required to effectively fine-tune such a model was significant, necessitating ongoing training and development for our team members. These challenges, while daunting, underscored the value of having a dedicated, specialized team focused on model development. Their expertise was key to overcoming the technical barriers presented by large-scale, custom AI model development. Data preprocessing efforts were also considerable, involving the cleaning, integration, and transformation of large datasets to ensure they were fit for use in a high-performance modeling environment.

Despite its outperformance in multiple benchmarks, KatzGPT's development journey underscores the complexities involved in custom model building within the LLM domain. The process revealed critical limitations of current LLMs, such as their struggle with out-of-domain queries. KatzGPT, while excelling in contextually accurate responses, demonstrated this limitation when handling inputs outside the embedding space of the training data, a known issue across many LLMs. This presents a key area for future research: developing more dynamic embedding techniques or expanding training datasets to cover a wider range of real-world scenarios. Moreover, the tendency of larger models to incorporate extraneous details into their responses sometimes affected KatzGPT's ROUGE scores negatively, highlighting the trade-off between model size and response precision. Balancing these factors remains a significant challenge, but it also offers opportunities for fine-tuning model architectures to achieve both breadth and accuracy in future iterations.

In conclusion, the development and refinement of KatzGPT have not only pushed the boundaries of what is achievable with custom LLMs but also provided critical insights into the relationship between data quality, model architecture, and fine-tuning strategies. The lessons learned from this project contribute to the academic community’s understanding of artificial intelligence and pave the way for future innovations in the field. As AI technology continues to evolve, our experience with KatzGPT emphasizes that a model's success is not solely defined by its scale but by the intelligent integration of data management, architecture optimization, and sequential fine-tuning strategies.

% =================================================================================

\section{Conclusion}\label{sec13}

In this paper, we developed KatzBot, a chatbot designed to improve academic communication through the introduction of KatzGPT, a custom-built large language model (LLM). Our model demonstrates state-of-the-art performance in generating contextually accurate and coherent responses, significantly enhancing text comprehension and response generation capabilities in university environments. The success of KatzGPT underscores the transformative potential of targeted model training strategies and the capabilities of custom LLMs tailored for specialized tasks.

The extensive experiment results demonstrate that KatzGPT consistently outperforms existing models across several key benchmarks. This performance is a direct result of our innovative approach to sequential fine-tuning, data preprocessing, and hyperparameter optimization, which allowed KatzGPT to achieve higher ROUGE scores compared to established models like GPT-2, Llama2, and Microsoft Phi series. The development of KatzGPT highlights the importance of investing in high-quality, diverse datasets and adopting a task-specific fine-tuning approach to achieve significant improvements in AI performance. Overall, this project illustrates the critical role that custom-built models can play in advancing academic AI applications, pushing the boundaries of what current LLMs can achieve.

\bmhead{Acknowledgements}

All authors contributed to the design and development of KatzGPT. Author 1 led the design, implementation, and training of the model architecture. Author 2 was responsible for data preprocessing, while Author 3 integrated voice recognition and multilingual support.

% =================================================================================

\end{document}